# Fuzzy Expert Systems for Prediction of ICU Admission in Patients with COVID-19

A. A. Sadat Asl[a,*], M. M. Ershadi[b,1], S. Sotudian[b,1], X. Li[a], S. Dick[a]

[a]Department of Electrical and Computer Engineering, University of Alberta, Edmonton, AB, Canada
[b]Department of Industrial Engineering, Amirkabir University of Technology, Tehran, Iran

## Abstract

The pandemic COVID-19 disease has had a dramatic impact on almost all countries around the world so that many hospitals have been overwhelmed with Covid-19 cases. As medical resources are limited, deciding on the proper allocation of these resources is a very crucial issue. Besides, uncertainty is a major factor that can affect decisions, especially in medical fields. To cope with this issue, we use fuzzy logic (FL) as one of the most suitable methods in modeling systems with high uncertainty and complexity. We intend to make use of the advantages of FL in decisions on cases that need to treat in ICU. In this study, an interval type-2 fuzzy expert system is proposed for prediction of ICU admission in COVID-19 patients. For this prediction task, we also developed an adaptive neuro-fuzzy inference system (ANFIS). Finally, the results of these fuzzy systems are compared to some well-known classification methods such as Naive Bayes (NB), Case-Based Reasoning (CBR), Decision Tree (DT), and K Nearest Neighbor (KNN). The results show that the type-2 fuzzy expert system and ANFIS models perform competitively in terms of accuracy and F-measure compared to the other system modeling techniques.

## 1. Introduction

COVID-19 or Coronavirus has affected the public health and economics of many countries in the world due to its contagious nature and lack of effective medicine or vaccine [1]. It has spread to over 193 million people worldwide by the end of July 2021. According to medical reports, the mortality rate associated with this virus is low. However, the long duration of the disease and the disability of patients for a long time cause further spread of the disease and thus increase the mortality associated with this disease. It has killed over 4 million people by the end of July 2021 [2].

It is noteworthy that many COVID-19 patients will develop mild to moderate illness and recover without hospitalization. Fever, dry cough, and tiredness are the most common symptoms of COVID-19. These patients are advised to manage their symptoms at home. On the other hand, serious symptoms such as difficulty breathing and chest pain have appeared in some patients during the disease [3]. This condition can quickly get worse so that an emergency situation occurs. These critical situations can increase human decision-making errors leading to more financial and non-financial losses.

Besides, predicting the situation of patients is useful for hospitals and health centers due to the spread of this virus. It can help to design targeted tests of people, predict the number of required resources in hospitals and health centers, and inform medical plans for prioritizing the level of care, design-related policy about vaccination, and so on. [4, 5]. In this way, health centers and hospitals can appropriately allocate their limited resources including treatment in intensive care units (ICUs), tests, and ventilators to the patients.

---

[1] The authors contributed equally to this paper.
* Corresponding author: A.A. Sadat Asl

Therefore, to prevent human decision-making errors and to properly allocate the limited resources, it is a rational idea to design an expert system for the prediction of resource utilization in patients with COVID-19. In this study, the COVID-19 patient pre-condition dataset provided by the Mexican government is used [6]. To the best of our knowledge, there are a few expert systems for the prediction of a problem related to the COVID-19 patients. However, most of them use direct approach in which rules are extracted by an expert. In this paper, our expert system uses fuzzy modeling with an indirect approach, where the rules are extracted automatically by implementing a clustering approach. Furthermore, for the first time, an expert system is designed for the prediction of ICU admission in COVID-19 patients. In addition to developing a type-2 fuzzy expert system, an adaptive neuro-fuzzy inference system and some classification methods such as NB, CBR, DT, and KNN are implemented. Finally, we compare the results of these techniques in terms of accuracy and F-measure.

The structure of this paper is as follows: a literature review of different expert systems for medical diagnosis is presented in section 2. The steps of designing the type-2 fuzzy expert system are presented in section 3. In section 4, the interval type-2 fuzzy expert system for ICU admission is presented. Classification methods and the developed fuzzy expert systems are evaluated in section 5. Finally, conclusions and some directions for future studies are considered in section 6.

## 2. Literature review

In this section, published research of Web of Science between the years 1980 and 2021 about the "COVID-19" or "Coronavirus" and "fuzzy expert systems" are analyzed using VOSviewer 1.6.10 software, and the corresponding results are presented in Figure 1. Different circles show a different number of articles and the larger circles are associated with more articles. The lines between circles show the relationships among their articles in terms of their references. Besides, VOSviewer 1.6.10 software clusters different circles into four colored clusters according to their relationships. Figure 2 shows the results according to the year of the published articles.

According to the published research of Web of Science, a search with "COVID-19" or "Coronavirus" or other equivalent terms found more than 1755000 results. Besides, "expert systems" is appeared in more than 1125000 papers. However, less than 100 paper present a fuzzy expert system for COVID-19 or Coronavirus diagnosis and no paper considers ICU admission in patients with COVID-19 using a fuzzy expert system. It demonstrates a serious gap due to the importance of the COVID-19 epidemic and the application of expert systems for medical decision support systems. Therefore, in this paper, we are going to fill this gap by designing fuzzy expert systems for the prediction of ICU admission in COVID-19 patients.

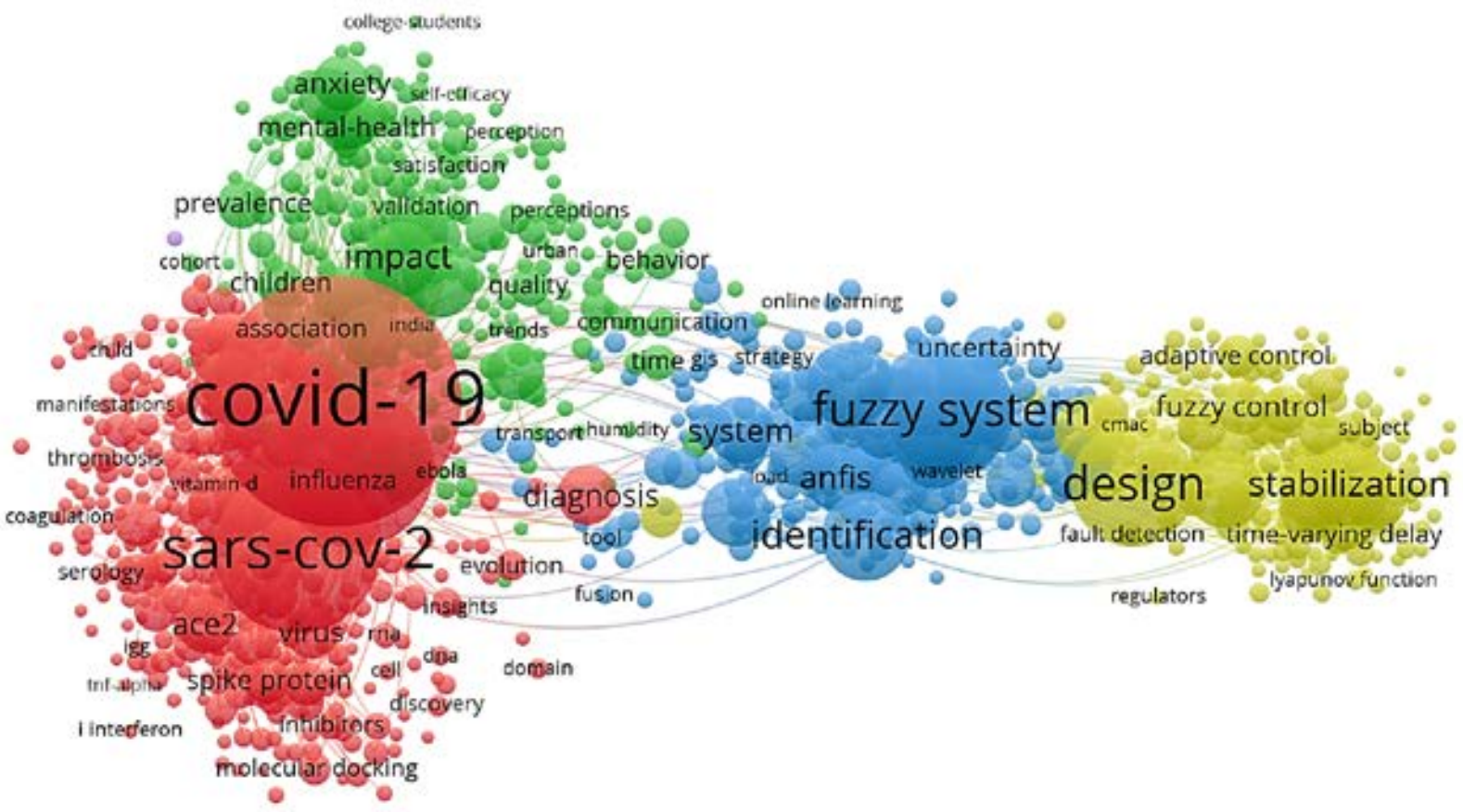

**Figure 1.** Different applications of COVID-19 and expert systems.

Thanks to the importance of expert systems, many researchers have used them in medical fields for the diagnosis of various diseases. For example, Hamedan et al. proposed a fuzzy expert system to predict chronic kidney disease. In their work, first, they identified the diagnostic parameters and risk factors through a literature review and a survey of some nephrologists, then a set of fuzzy rules for the prediction of chronic kidney disease was determined [26]. In another study, Hussain et al. presented a multi-layered fuzzy Mamdani inference system to analyze the prevailing thyroid disease. In their study, the proposed expert system is based on two layers. In layer 1, the presence or absence of thyroid disease is diagnosed. If layer 1 indicates the presence of thyroid disease, then layer 2 is activated by which the type of thyroid disease is determined [27].

Khalil et al. developed a new fuzzy soft expert system to predict lung cancer disease by using weight loss, shortness of breath, chest pain, persistent cough, blood in sputum, and age of patients. In their work, a prediction of the fuzzy soft expert system is composed of four main steps: 1) Transforming real-valued inputs into fuzzy numbers. 2) Converting fuzzy numbers into fuzzy soft sets. 3) Reducing the family of fuzzy soft sets obtained to a new family of fuzzy soft sets. 4) Using the proposed method to get the output data [29]. Mahanta and Panda developed a fuzzy expert system for the prediction of prostate cancer. In their study, age, prostate-specific antigen (PSA), prostate volume (PV), and Free PSA (FPSA) are fed as inputs into the system, and prostate cancer risk (PCR) is obtained as the output [30].

Mirmozaffari et al. presented an expert system for diagnosing the type of gastrointestinal disease and determining the type of tests needed to diagnose the disease [31]. Mojrian et al. presented a method based on a multilayer fuzzy expert system for the detection of breast cancer using an extreme learning machine (ELM) classification model integrated with radial basis function (RBF) kernel called ELM-RBF. In this study, they showed that their proposed method outperforms the linear-SVM model [32].

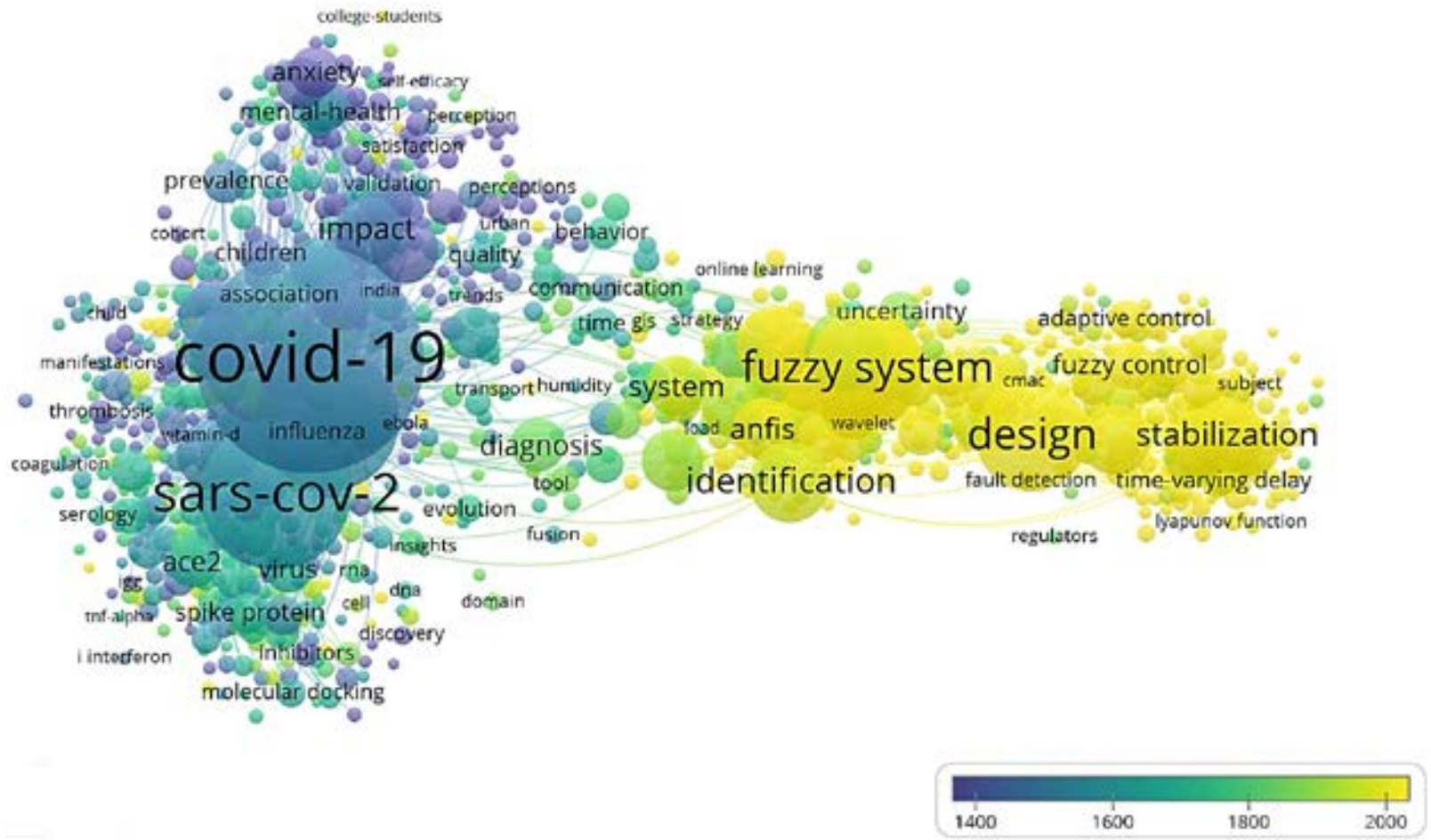

**Figure 2.** The year of published articles about COVID-19 and expert systems.

In another study, Naseer et al. proposed a fuzzy expert system for the diagnosis of heart disease. In their proposed system, the input variables comprised of age, chest pain, electrocardiography, blood pressure systolic, diabetes, and cholesterol are transmitted with the help of fuzzy rules which are framed in the light of low, normal, high, and very high intensity and the output is obtained using the Mamdani Inference method diagnosing the heart disease [33]. Siddiqui et al. developed an adaptive hierarchical Mamdani fuzzy expert system for the detection of arthritis. In their research, the expert system is comprised of two layers. In the first layer, the input variables are rest pain, morning stiffness, body pain, joint infection, swelling, redness, past injury and age that detects output condition of arthritis to be normal, infection and/or other problem, and in the second layer, the type of arthritis is diagnosed [34]. In the above studies, the authors designed various fuzzy expert systems in different medical fields.

However, there are a few studies on coronavirus-related issues, most of which are based on a direct approach where the rules are obtained by experts. Our expert system in this research employs fuzzy modeling using an indirect method, in which the rules are extracted automatically using a clustering approach. We develop an expert system designed specifically for the prediction of ICU admission in COVID-19 patients. Table 1 summarizes the fuzzy expert systems proposed for the diagnosis of various types of diseases.

**Table 1.** A review on the application of fuzzy expert systems in medical fields.

| Author [year] | Disease | Technique |
|---|---|---|
| Sotudian et al. (2016) [7] | Hepatitis | Indirect approach for fuzzy system modeling |
| Guzmán et al. (2017) [8] | Blood pressure | Neuro-fuzzy hybrid model |
| Meza-Palacios et al. (2017) [9] | Type-2 diabetes mellitus | Fuzzy expert system based on experts' guidelines |
| Sadat Asl & Zarandi (2017) [10] | Leukemia | Mamdani-style Type-2 fuzzy expert system |
| Zarandi et al. (2017) [11] | Heart disease | Expert system based on fuzzy bayesian network |
| Motlagh et al. (2018) [12] | Depressive disorder | Web-based fuzzy expert system |
| Caliwag et al. (2018) [13] | Venereal diseases | Mobile expert system based on fuzzy logic |
| Muhammad et al. (2018) [14] | Coronary artery disease | Expert system with developed knowledge acquisition |
| Nazari et al. (2018) [15] | Heart diseases | Fuzzy inference-fuzzy analytic hierarchy process-based |
| Soltani et al. (2018) [16] | Glaucoma | Expert fuzzy logic & image processing methods |
| Tuan et al. (2018) [17] | Dental disorders | Fuzzy computing & image processing methods |
| Terrada et al. (2018) [18] | Cardiovascular diseases | Fuzzy medical system using risk factors based on data |
| Ahmad et al. (2019) [19] | Hepatitis B | Multilayer Mamdani fuzzy inference system |
| Raza et al. (2019) [20] | Erythemato squamous | Expert system based on fuzzy rules |
| Arji et al. (2019) [21] | Infectious disease | Fuzzy inference & rule-based fuzzy logic, ANFIS |
| Kaur & Kakkar (2019) [22] | Neurodevelopmental disorders | Fuzzy based-systems based on co-morbid factors |
| Sadat Asl (2019) [23] | Leukemia | Two-stage expert system based on type-2 fuzzy logic |
| Mirmozaffari (2019) [24] | Liver diseases | Expert system based on the VP-Expert shell |
| Mujawar & Jadhav (2019) [25] | Diabetes | Web-based fuzzy expert system |
| Mutawa and Alzuwawi (2019) [26] | Uveitis | Multilayered rule-based expert system |
| Sajadi et al. (2019) [27] | Hypothyroidism | Fuzzy rule-based expert system |
| Ershadi et al. (2020) [28] | Different diseases and cancers | Multi-classifier based on fuzzy expert system |
| Hamedan et al. (2020) [29] | Chronic kidney disease | Fuzzy expert system with Mamdani inference system |
| Hussain et al. (2020) [30] | Thyroid disease | Multi-layered fuzzy Mamdani inference system |
| Khalil et al. (2020) [31] | Lung cancer | Fuzzy soft expert system |
| Mahanta & Panda (2020) [32] | Prostate cancer | Fuzzy expert system with Mamdani inference system |
| Mirmozaffari (2020) [33] | Gastrointestinal diseases | Fuzzy expert system based on VP-Expert shell |
| Mojrian et al. (2020) [34] | Breast cancer | Multilayer fuzzy expert system based on RBF |
| Naseer et al. (2020) [35] | Heart disease | Mamdani fuzzy inference expert system |
| Siddiqui et al. (2020) [36] | Arthritis | Adaptive hierarchical Mamdani fuzzy expert system |

## 3. Designing the type-2 fuzzy expert system

Artificial intelligence (AI) has been extensively applied in various fields of science, such as engineering and medical sciences [37, 38, 39, 40]. Expert systems are one of the most commercially successful branches of AI. They are programs with a broad base of knowledge emulating expert problem-solving skills in a certain domain [41]. Besides, fuzzy logic is a suitable modeling method for systems with high uncertainty and complexity by which a framework can be provided for reasoning, inference, control, and decision making [42]. Due to common features and others that complement each other, fuzzy logic and expert system technologies can easily be integrated. Both techniques are well-suited to deal with decision-making and other knowledge-oriented problems. Systems with these techniques have enhanced efficiency, system quality, and speed of execution [43].

As mentioned earlier, the main objective of this paper is to design an expert system to predict ICU needs for COVID-19 patients based on type-2 FL. To this end, there are two common approaches for the selection of the parameters of a type-2 FL system. The first one is the partially dependent approach. In this approach, first, the best possible type-1 FL system is designed. Then, it is used to initialize the parameters of a type-2 FL system. The second one is the totally independent approach where all of the parameters of the type-2 FL system are tuned without using an existing type-1 design [44]. In this paper, we use the partially dependent approach because of its advantages compared with the totally dependent approach. After designing a type 1 fuzzy system, a type-2 rule-based fuzzy system with uncertain standard deviation and interval-valued membership function is implemented. The same rules of the type-1 fuzzy system are used by this system and the only difference is that if-part and then-part are type-2. The designed system is comprised of the following steps:

- Data preprocessing and determining the inputs and output of the system;
- Clustering the output space and determination of the number of rules;
- Projection of membership functions of the output onto the inputs to obtain the membership values of the inputs;
- Tuning the parameters of type-1 membership functions of inputs and output variables;
- Transforming type-1 to interval type-2 membership functions;
- Tuning the parameters of type-2 membership functions.

### 3.1. Data Preprocessing

We use a publicly available dataset, containing information of about 566602 patients. The data contains some information about pregnancy, diabetes, chronic obstructive pulmonary disease (COPD), asthma, cardiovascular, obesity, tobacco, etc., and determines whether a patient needs an ICU or not [6]. In this dataset, there are 444814 missing data for ICU. Therefore, the final number of patients used in fuzzy expert systems and other classification techniques is 121788. Besides, a few outliers (e.g., male patient's pregnancy) were detected and removed from the dataset.

Most precondition features of this dataset are categorical, each of which takes the value yes, no, and unspecified. These categorical features were converted to numerical by ‘one-hot’ encoding. By creating auxiliary variables that help differentiate between various categories of a feature, one-hot encoding transforms the categorical feature into multiple binary variables. Finally, highly correlated variables were eliminated as they provide similar information. Specifically, we calculated pairwise correlations among the variables, and one among any two highly correlated variables with an absolute correlation coefficient higher than 0.85 was removed. After the preparation of the dataset, we reached 27 features and one output determines whether a patient needs to the ICU or not.

### 3.2. Clustering the output space and determination of the number of rules

To determine the number of rules, we use Fukuyama cluster validity index which can be defined as follows [45]: (1)

$$\min_{2\le c\le C_{max}} V_{FS}(U,V,X) = \sum_{i=1}^{c}\sum_{j=1}^{N} u_{ij}^{m}\left\|x_j - v_i\right\|^2 - \sum_{i=1}^{c}\sum_{j=1}^{N} u_{ij}^{m}\|v_i - \bar{v}\|^2, \tag{1}$$

where $X = \{x_1, x_2, \dots, x_N\} \subseteq \mathbb{R}^d$ is the dataset in d-dimensional vector space, $u_{ij}$ is the degree of belonging of the $j^{th}$ data to the $i^{th}$cluster, $V = \{v_1, v_2, \dots, v_c\}$ is the prototypes of clusters, c is the number of clusters, $\bar{v} = \frac{\sum v_i}{c}$, m is the degree of fuzziness, U is fuzzy partition matrix, and N is the number of samples. By solving $\min_{2 \le c \le C_{max}} V_{FS}$, the optimal cluster number is obtained. In this study, this cluster validity index was implemented and the optimal value for the number of clusters is obtained five clusters. Therefore, we have five rules in our system. In the proposed system, Mamdani inference system is used where the antecedents and consequents of the rule-based system are fuzzy sets. We clustered the output data and then obtained the output clusters' primary membership grades using Sugeno and Yasukawa method [46]. First, we partition the output space, and then, get the input space clusters by projecting the output space partition to each input variable space.

### 3.3. Projection of membership functions of the output onto the inputs

For the input variables, the appropriate membership grades should be calculated after clustering the output space. One way is to set each input's membership grade equal to its corresponding output membership grade obtained by the procedure of output data clustering. In this way, for each output data, all the related input variables would then have a similar membership grade. The issue with this approach is that the membership functions are not convex and a further approximation is required to form the convex membership functions. Furthermore, the output membership grade is not always the same as the input membership grades at each sample point [47]. For these reasons, in this paper, Zarandi's approach is used. According to this approach, first, the ranges in which the membership functions of the input variable adopt value 1 are determined. The data points are then classified using the GK method, by given m and c determined in the preceding step (obtained from output variable clustering stage) and analyzing the objective function of classification algorithm (for more details please refer to [47]).

#### 3.4. Tuning the parameters of type-1 membership functions

There are several parameters in type-1 FL systems that can either be pre-specified or can be tuned during a training phase. An impeccable FL system should have $f(x) = d$, where d is the desired output. However, there are typically errors between the desired and actual output. Therefore, in order to produce better results, tuning the parameters of the fuzzy model is important. In this paper, the suggested tuning algorithm by Liang and Mendel is used. Based on this approach, all of the parameters related to a Gaussian type-1 are tuned using the steepest descent method. Given an input-output training pair $\left(x^{(i)}, y^{(i)}\right), x^{(i)} \in R^G$ and $y^{(i)} \epsilon R$, a type-1 fuzzy is designed by minimizing the following error function [48]:

$$e(t) = \frac{1}{2}[f(x^{(i)}) - y^{(i)}]^2, \qquad i = 1, \dots, N. \tag{2}$$

#### 3.5. Transforming type-1 to interval type-2 membership functions

To transform type-1 to an interval type-2 fuzzy set with uncertain standard deviation, the case of a Gaussian primary membership function with a fixed mean $m_f^S$ and uncertain standard deviation that takes on values in $[\sigma_{f_1}^S, \sigma_{f_2}^S]$ is considered [48]:

$$u_f^S(x_f) = exp\left[-\frac{1}{2}\left(\frac{x_f - m_f^S}{\sigma_f^S}\right)\right], \qquad \sigma_f^S \in [\sigma_{f_1}^S, \sigma_{f_2}^S], \tag{3}$$

where $f = 1, \ldots, G$ ; $G$ is the number of antecedents; $S = 1, \ldots, D$ ; and $D$ is the number of rules. We can obtain the upper and lower membership functions by replacing $\sigma_{f_2}^S$ and $\sigma_{f_1}^S$ with $\sigma_f^S$ in the above expression.

### 3.6. Tuning the parameters of type-2 membership functions

For tuning the parameters of the interval type-2 FL system, we use the proposed tuning algorithm by Liang and Mendel. In interval type-2 system, $f(x)$ is determined by upper and lower membership functions and centroids of interval type-2 fuzzy sets, and therefore, we want to tune these parameters. Since an interval type-2 FL system can be identified by two fuzzy basis function expansions, we can focus on tuning the parameters of just these two type-1 FL systems [49].

### 4. Proposed interval type-2 fuzzy expert system

In this paper, we use Mamdani inference system in which the antecedents and consequent are type-2 fuzzy sets that have a fixed mean and an uncertain standard deviation that takes values in an interval. The interval type-2 FL system is created from the type-1 FL system. The proposed system uses singleton fuzzification, product t-norm, product inference, and center-of-sets type-reduction, with the same number of fuzzy sets and rules as the type-1 FL system. Several defuzzification methods have been used, such as centroid, bisector, and Yager. The best results of this system are obtained by Yager defuzzification method. Figure 3 demonstrates the rule-based and inference mechanism for the proposed interval type-2 fuzzy system. Table 2 shows both consequent and antecedent parameters of the expert system. In this table, $\mu$, $\overline{\sigma}$ and $\underline{\sigma}$ are the fixed mean, upper bound standard deviation and lower bound standard deviation, respectively.

## 5. Performance Evaluation and Analysis

In this paper, in addition to developing a type-2 fuzzy expert system for the prediction of ICU admission, we develop the ANFIS model for this prediction task using MATLAB toolbox. Then, we compare the performance of the developed fuzzy expert systems to several well-known classification methods such as NB, CBR, DT, and KNN in terms of accuracy and F-measure. To evaluate the performance of each system, the dataset is divided into training and test sets. In this way, for each system modeling technique, 70% of the dataset has been used for training. Table 3 shows the accuracy and F-measure of different system modeling techniques implemented in this study.

Classification accuracy is the total number of accurate predictions divided by the total number of predictions made for a dataset. As a performance measure, accuracy is not sufficient for imbalanced classification problems. The key explanation is that the vast number of examples from the dominant class or classes would overwhelm the number of minority class examples. Using precision and recall metrics is an alternative to using classification accuracy. F-Measure integrates both precision and recall into a single measure and captures both properties [50]. As can be seen in Table 3, the developed fuzzy expert systems could achieve an accuracy of 91.6% and an F-measure of 95.6%. Comparing to the CBR and KNN methods, the fuzzy expert systems can improve the accuracy and F-measure by about 1.5% to 5% and 1% to 3%, respectively. Furthermore, the developed fuzzy models performed competitively compared to the other classification methods. Overall, the results show that both type-2 fuzzy system and ANFIS model could outperform the well-known classification methods.

**Table 2.** The antecedent and consequent parameters of the system.

| | **Parameters** | **Rule 1** | **Rule 2** | **Rule 3** | **Rule 4** | **Rule 5** |
|---|---|---|---|---|---|---|
| Antecedent-Var 1 | $\boldsymbol{\mu}$ | 32.562 | 79.019 | 63.131 | 49.029 | 4.849 |
| | $\overline{\boldsymbol{\sigma}}$ | 9.501 | 10.240 | 7.552 | 7.319 | 15.489 |
| | $\underline{\boldsymbol{\sigma}}$ | 5.700 | 6.144 | 4.531 | 4.391 | 9.294 |
| Antecedent-Var 2 | $\boldsymbol{\mu}$ | 1.578 | 1.575 | 1.608 | 1.632 | 1.562 |
| | $\overline{\boldsymbol{\sigma}}$ | 0.362 | 0.357 | 0.444 | 0.440 | 0.253 |
| | $\underline{\boldsymbol{\sigma}}$ | 0.254 | 0.250 | 0.311 | 0.308 | 0.177 |
| Antecedent-Var 3 | $\boldsymbol{\mu}$ | 1.509 | 1.650 | 1.656 | 1.626 | 1.408 |
| | $\overline{\boldsymbol{\sigma}}$ | 0.372 | 0.337 | 0.425 | 0.442 | 0.260 |
| | $\underline{\boldsymbol{\sigma}}$ | 0.186 | 0.169 | 0.212 | 0.221 | 0.130 |
| Antecedent-Var 4 | $\boldsymbol{\mu}$ | 1.004 | 1.005 | 1.005 | 1.006 | 1.002 |
| | $\overline{\boldsymbol{\sigma}}$ | 0.046 | 0.053 | 0.058 | 0.076 | 0.023 |
| | $\underline{\boldsymbol{\sigma}}$ | 0.041 | 0.047 | 0.052 | 0.068 | 0.021 |
| Antecedent-Var 5 | $\boldsymbol{\mu}$ | 1.102 | 1.376 | 1.406 | 1.284 | 1.014 |
| | $\overline{\boldsymbol{\sigma}}$ | 0.186 | 0.337 | 0.441 | 0.381 | 0.095 |
| | $\underline{\boldsymbol{\sigma}}$ | 0.130 | 0.236 | 0.308 | 0.267 | 0.066 |
| Antecedent-Var 6 | $\boldsymbol{\mu}$ | 1.004 | 1.005 | 1.004 | 1.005 | 1.002 |
| | $\overline{\boldsymbol{\sigma}}$ | 0.042 | 0.050 | 0.050 | 0.060 | 0.019 |
| | $\underline{\boldsymbol{\sigma}}$ | 0.034 | 0.040 | 0.040 | 0.048 | 0.015 |
| Antecedent-Var 7 | $\boldsymbol{\mu}$ | 1.007 | 1.127 | 1.053 | 1.020 | 1.002 |
| | $\overline{\boldsymbol{\sigma}}$ | 0.023 | 0.139 | 0.089 | 0.045 | 0.015 |
| | $\underline{\boldsymbol{\sigma}}$ | 0.014 | 0.083 | 0.053 | 0.027 | 0.009 |
| Antecedent-Var 8 | $\boldsymbol{\mu}$ | 1.033 | 1.021 | 1.022 | 1.025 | 1.032 |
| | $\overline{\boldsymbol{\sigma}}$ | 0.044 | 0.031 | 0.040 | 0.045 | 0.030 |
| | $\underline{\boldsymbol{\sigma}}$ | 0.030 | 0.022 | 0.028 | 0.032 | 0.021 |
| Antecedent-Var 9 | $\boldsymbol{\mu}$ | 1.005 | 1.006 | 1.005 | 1.006 | 1.004 |
| | $\overline{\boldsymbol{\sigma}}$ | 0.048 | 0.058 | 0.061 | 0.071 | 0.030 |
| | $\underline{\boldsymbol{\sigma}}$ | 0.038 | 0.046 | 0.049 | 0.057 | 0.024 |
| Antecedent-Var 10 | $\boldsymbol{\mu}$ | 1.040 | 1.034 | 1.034 | 1.030 | 1.109 |
| | $\overline{\boldsymbol{\sigma}}$ | 0.056 | 0.049 | 0.062 | 0.057 | 0.083 |
| | $\underline{\boldsymbol{\sigma}}$ | 0.033 | 0.029 | 0.037 | 0.034 | 0.050 |
| Antecedent-Var 11 | $\boldsymbol{\mu}$ | 1.004 | 1.005 | 1.004 | 1.006 | 1.002 |
| | $\overline{\boldsymbol{\sigma}}$ | 0.044 | 0.049 | 0.052 | 0.069 | 0.019 |
| | $\underline{\boldsymbol{\sigma}}$ | 0.026 | 0.030 | 0.031 | 0.041 | 0.012 |
| Antecedent-Var 12 | $\boldsymbol{\mu}$ | 1.111 | 1.552 | 1.443 | 1.277 | 1.014 |
| | $\overline{\boldsymbol{\sigma}}$ | 0.202 | 0.370 | 0.454 | 0.382 | 0.107 |
| | $\underline{\boldsymbol{\sigma}}$ | 0.141 | 0.259 | 0.318 | 0.267 | 0.075 |
| Antecedent-Var 13 | $\boldsymbol{\mu}$ | 1.007 | 1.009 | 1.007 | 1.008 | 1.013 |
| | $\overline{\boldsymbol{\sigma}}$ | 0.076 | 0.085 | 0.094 | 0.104 | 0.083 |
| | $\underline{\boldsymbol{\sigma}}$ | 0.061 | 0.068 | 0.075 | 0.083 | 0.066 |
| Antecedent-Var 14 | $\boldsymbol{\mu}$ | 1.050 | 1.072 | 1.054 | 1.044 | 1.151 |
| | $\overline{\boldsymbol{\sigma}}$ | 0.072 | 0.093 | 0.095 | 0.083 | 0.111 |
| | $\underline{\boldsymbol{\sigma}}$ | 0.043 | 0.056 | 0.057 | 0.050 | 0.067 |

**Table 2. (continued)** The antecedent and consequent parameters of the system.

| | Parameters | Rule 1 | Rule 2 | Rule 3 | Rule 4 | Rule 5 |
|---|---|---|---|---|---|---|
| Antecedent-Var 15 | $\mu$ | 1.016 | 1.116 | 1.058 | 1.028 | 1.046 |
| | $\overline{\sigma}$ | 0.035 | 0.132 | 0.098 | 0.058 | 0.046 |
| | $\underline{\sigma}$ | 0.024 | 0.092 | 0.068 | 0.041 | 0.032 |
| Antecedent-Var 16 | $\mu$ | 1.005 | 1.005 | 1.004 | 1.006 | 1.002 |
| | $\overline{\sigma}$ | 0.048 | 0.049 | 0.058 | 0.074 | 0.019 |
| | $\underline{\sigma}$ | 0.038 | 0.039 | 0.046 | 0.059 | 0.015 |
| Antecedent-Var 17 | $\mu$ | 1.199 | 1.153 | 1.218 | 1.255 | 1.026 |
| | $\overline{\sigma}$ | 0.237 | 0.201 | 0.312 | 0.346 | 0.078 |
| | $\underline{\sigma}$ | 0.166 | 0.141 | 0.218 | 0.242 | 0.055 |
| Antecedent-Var 18 | $\mu$ | 1.004 | 1.005 | 1.004 | 1.005 | 1.002 |
| | $\overline{\sigma}$ | 0.044 | 0.051 | 0.051 | 0.065 | 0.021 |
| | $\underline{\sigma}$ | 0.026 | 0.030 | 0.030 | 0.039 | 0.012 |
| Antecedent-Var 19 | $\mu$ | 1.049 | 1.070 | 1.074 | 1.048 | 1.017 |
| | $\overline{\sigma}$ | 0.071 | 0.091 | 0.121 | 0.088 | 0.029 |
| | $\underline{\sigma}$ | 0.042 | 0.055 | 0.072 | 0.053 | 0.017 |
| Antecedent-Var 20 | $\mu$ | 1.005 | 1.005 | 1.004 | 1.006 | 1.002 |
| | $\overline{\sigma}$ | 0.046 | 0.050 | 0.056 | 0.070 | 0.020 |
| | $\underline{\sigma}$ | 0.028 | 0.030 | 0.034 | 0.042 | 0.012 |
| Antecedent-Var 21 | $\mu$ | 1.089 | 1.109 | 1.092 | 1.083 | 1.008 |
| | $\overline{\sigma}$ | 0.119 | 0.134 | 0.151 | 0.142 | 0.032 |
| | $\underline{\sigma}$ | 0.083 | 0.094 | 0.106 | 0.100 | 0.023 |
| Antecedent-Var 22 | $\mu$ | 1.456 | 1.572 | 1.545 | 1.497 | 1.404 |
| | $\overline{\sigma}$ | 0.372 | 0.362 | 0.463 | 0.470 | 0.255 |
| | $\underline{\sigma}$ | 0.297 | 0.289 | 0.370 | 0.376 | 0.204 |
| Antecedent-Var 23 | $\mu$ | 1.202 | 1.093 | 1.117 | 1.152 | 1.153 |
| | $\overline{\sigma}$ | 0.228 | 0.135 | 0.196 | 0.237 | 0.129 |
| | $\underline{\sigma}$ | 0.137 | 0.081 | 0.118 | 0.142 | 0.077 |
| Antecedent-Var 24 | $\mu$ | 1.384 | 1.335 | 1.258 | 1.258 | 1.746 |
| | $\overline{\sigma}$ | 0.347 | 0.323 | 0.363 | 0.368 | 0.257 |
| | $\underline{\sigma}$ | 0.243 | 0.226 | 0.254 | 0.258 | 0.180 |
| Antecedent-Var 25 | $\mu$ | 1.504 | 1.539 | 1.619 | 1.627 | 1.152 |
| | $\overline{\sigma}$ | 0.373 | 0.363 | 0.442 | 0.445 | 0.218 |
| | $\underline{\sigma}$ | 0.335 | 0.327 | 0.398 | 0.400 | 0.196 |
| Antecedent-Var 26 | $\mu$ | 1.002 | 1.001 | 1.001 | 1.001 | 1.003 |
| | $\overline{\sigma}$ | 0.021 | 0.013 | 0.012 | 0.019 | 0.019 |
| | $\underline{\sigma}$ | 0.017 | 0.010 | 0.010 | 0.015 | 0.015 |
| Antecedent-Var 27 | $\mu$ | 1.041 | 1.000 | 1.000 | 1.001 | 1.008 |
| | $\overline{\sigma}$ | 0.326 | 0.021 | 0.023 | 0.034 | 0.057 |
| | $\underline{\sigma}$ | 0.228 | 0.015 | 0.016 | 0.024 | 0.040 |
| Consequent | $\mu$ | 1.066 | 1.075 | 1.082 | 1.080 | 1.164 |
| | $\overline{\sigma}$ | 0.095 | 0.102 | 0.138 | 0.136 | 0.121 |
| | $\underline{\sigma}$ | 0.057 | 0.061 | 0.083 | 0.082 | 0.072 |

Type-2 FL systems usually give better results than their type-1 counterparts as the type-2 fuzzy sets and systems generalize type-1 fuzzy sets and systems so that more uncertainty can be handled. However, in our case, as can be seen, the results of the type-2 fuzzy expert system are very close to the ANFIS model. In the dataset, as most input variables are categorical, the type-2 fuzzy set lost its efficiency. Therefore, in this specific problem, type-2 FL has lost its superiority over the developed ANFIS model.

**Table 3.** Accuracy and F-measure of different system modeling techniques.

| Methods | Accuracy % | F-measure % |
|---|---|---|
| **Type-2 fuzzy system** | 91.64 | 95.64 |
| **ANFIS** | 91.66 | 95.66 |
| **NB** | 90.79 | 95.15 |
| **CBR** | 86.26 | 92.54 |
| **DT** | 90.63 | 95.06 |
| **KNN** | 90.04 | 94.73 |

## 6. Conclusion

In this study, a type-2 fuzzy expert system and an adaptive neuro-fuzzy inference system are developed for the prediction of ICU admission. Furthermore, to evaluate the performance of these fuzzy systems, several classification methods such as NB, CBR, DT, and KNN are also implemented. All these methods are tested on a publicly available dataset. The results demonstrate the efficacy of the proposed fuzzy expert systems, with an accuracy of 91.6% and an F-measure of 95.6%, which outperform the other conventional classification techniques. When comparing the two developed fuzzy expert systems, we can notice that the results of the type-2 fuzzy expert system are very close to the ANFIS model since, in our problem, most input variables are categorical.

One of the limitations of the present study is a lack of publicly available COVID-19 datasets with detailed symptoms. Most features of the dataset used in this study are categorical which may affect the general superiority of type-2 expert system over the developed ANFIS model. Over time, more COVID-19 datasets will be published because more and more experiments are performed every day for this disease around the world. If the number of continuous input variables in new datasets is higher, the type-2 fuzzy expert system may provide better results due to its higher generalizability. Therefore, for future studies, the proposed fuzzy expert systems can be designed using some other datasets. A prominent feature of type-2 fuzzy logic is its resilience to noise compared to type-1 fuzzy logic. This resilience can be improved by introducing general type-2 fuzzy logic. Thus, an interesting future direction is to extend the proposed model to a general type-2 fuzzy expert system. Finally, we can extend the proposed model to deal with more complex data such as data with missing entries.

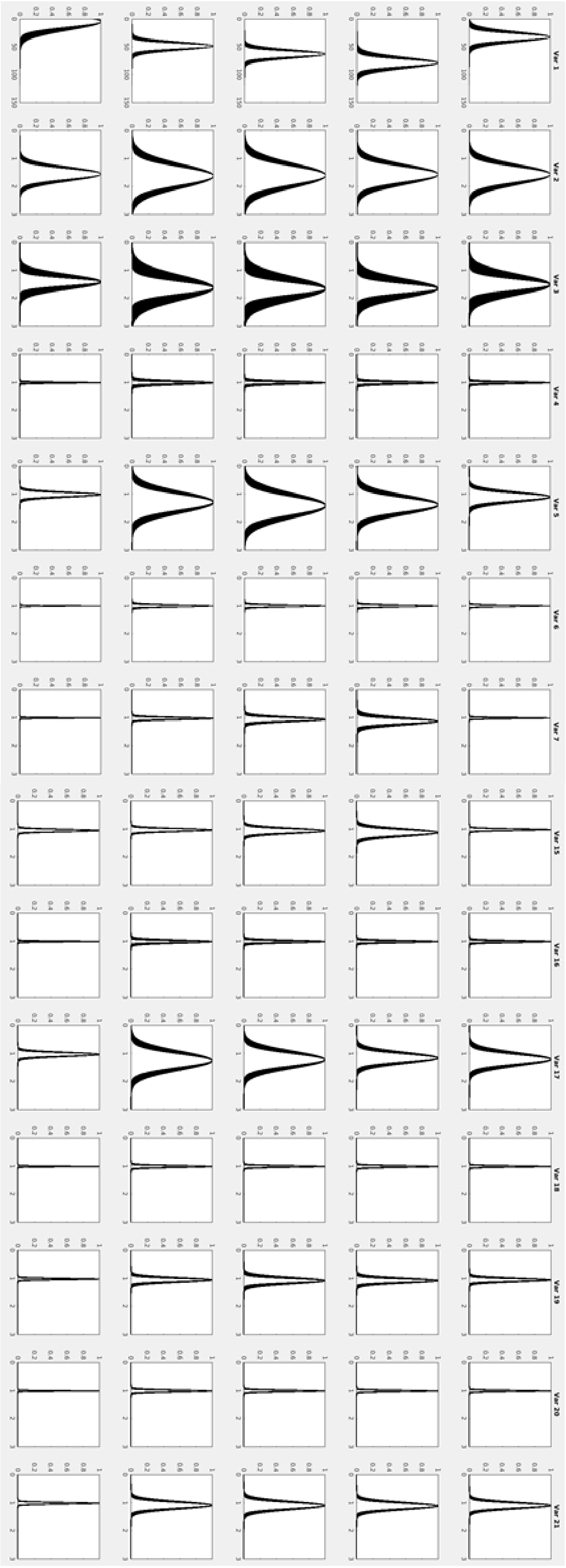

**Figure 3.** Interval type-2 fuzzy rule base.

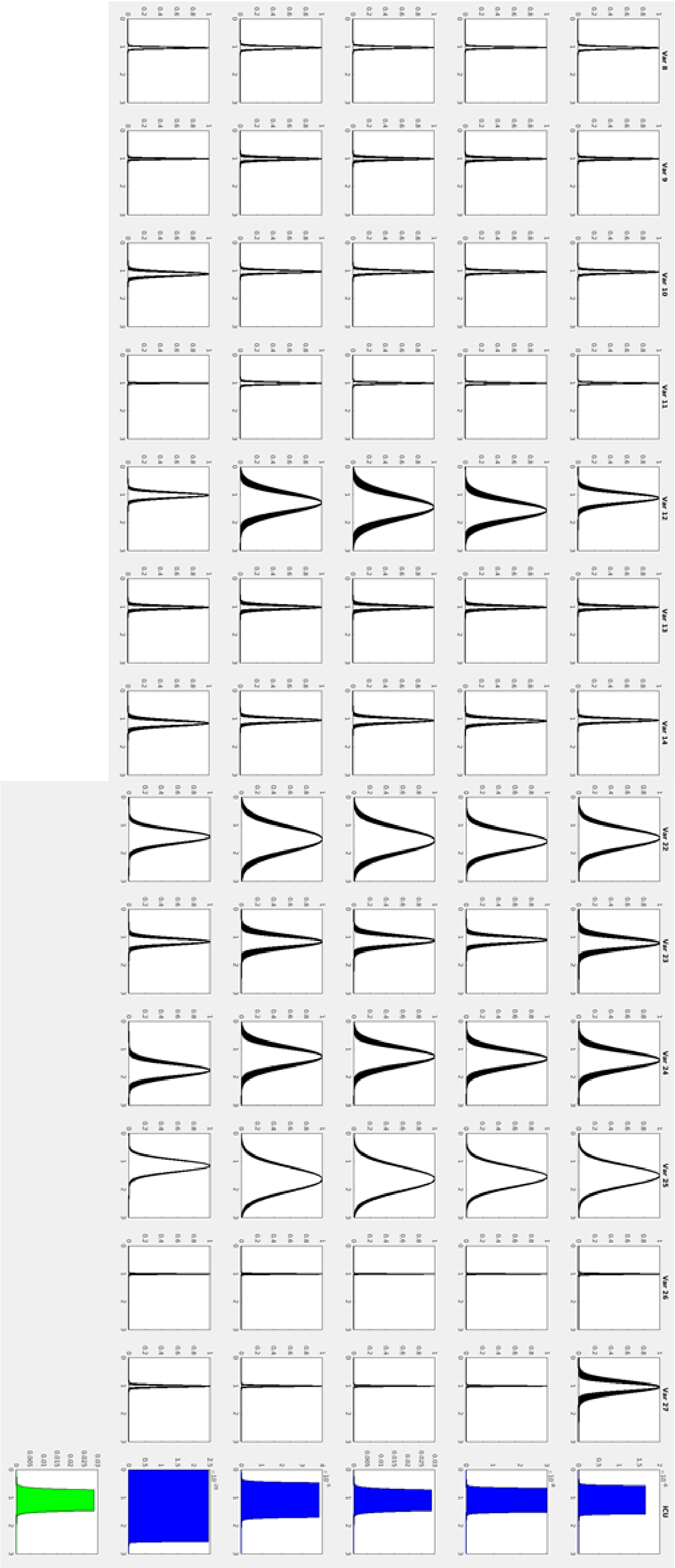

**Figure 3.** (continued) Interval type-2 fuzzy rule base.

Summary points:

| What is known? | What this adds? |
|---|---|
| • Hospitals have many problems with Covid-19 cases due to limited available medical resources. | • Proposed classifiers are designed in this paper to predict ICU admission in COVID-19 patients. |
| • There are high uncertainty and complexity to find the proper level of Covid-19 patients. | • Fuzzy logic methods are considered in this paper to model related uncertainty and complexity. |
| • The number of available data has significant effects on the learning phase of most classifiers. | • Different clinical features of 566602 COVID-19 patients are utilized in the learning phase of classifiers. |
| • There are different classifiers with special advantages and disadvantages. | • We compare the accuracies and F-measures of Naive Bayes (NB), Case-Based Reasoning (CBR), Decision Tree (DT), and K Nearest Neighbor (KNN) to the proposed type-2 fuzzy expert system and ANFIS model. |